\documentclass{article}
\usepackage{arxiv}

\usepackage[utf8]{inputenc} % allow utf-8 input
\usepackage[T1]{fontenc}    % use 8-bit T1 fonts
\usepackage{hyperref}       % hyperlinks
\usepackage{url}            % simple URL typesetting
\usepackage{booktabs}       % professional-quality tables
\usepackage{amsfonts}       % blackboard math symbols
\usepackage{nicefrac}       % compact symbols for 1/2, etc.
\usepackage{microtype}      % microtypography
\usepackage{lipsum}
\usepackage{graphicx}
\graphicspath{ {./images/} }
\usepackage{booktabs}  % Add this in your preamble if not already included

\usepackage[round]{natbib} 
\usepackage{xcolor}
\usepackage{caption}
\usepackage{hyperref}
\usepackage{cleveref}
\crefname{figure}{Fig.}{Figs.}
\hypersetup{
    colorlinks=true,
    linkcolor=blue,    % for \ref, \pageref, TOC
    citecolor=blue,    % for \cite
    urlcolor=blue      % for \url
}

\newcommand{\RN}[1]{\uppercase\expandafter{\romannumeral#1}}

\title{A Large Language Model-Empowered Agent for Reliable and Robust Structural Analysis}

\author{
Jiachen Liu$^{1,*}$,
Ziheng Geng$^{2,*}$,
Ran Cao$^{3}$,
Lu Cheng$^{4}$,
Paolo Bocchini$^{5}$,
Minghui Cheng$^{2,6\dagger}$\\
\\
$^{1}$Department of Electrical and Computer Engineering, University of Miami, Coral Gables, FL 33146, USA\\
$^{2}$Department of Civil and Architectural Engineering, University of Miami, Coral Gables, FL 33146, USA\\
$^{3}$College of Civil Engineering, Hunan University, Changsha, 410082, China\\
$^{4}$Department of Computer Science, University of Illinois Chicago, Chicago, IL 60607, USA\\
$^{5}$Center for Catastrophe Modeling and Resilience, Lehigh University, Bethlehem, PA 18015, USA\\
$^{6}$School of Architecture, University of Miami, Coral Gables, FL 33146, USA\\
\\
$^{*}$Equal contribution.\\
$^{\dagger}$Corresponding author: \texttt{minghui.cheng@miami.edu}
}

\begin{document}
\maketitle
\begin{abstract}
Large language models (LLMs) have exhibited remarkable capabilities across diverse open-domain tasks, yet their application in specialized domains such as civil engineering remains largely unexplored. This paper starts bridging this gap by evaluating and enhancing the reliability and robustness of LLMs in structural analysis of beams. Reliability is assessed through the accuracy of correct outputs under repetitive runs of the same analysis problems, whereas robustness is evaluated based on the performance across varying load and boundary conditions. A benchmark dataset, comprising eight beam structural analysis problems, is created to test the Llama-3.3 70B Instruct model. Results show that, despite an apparent qualitative understanding of structural mechanics, the LLM lacks the quantitative reliability and robustness required for engineering applications. To address these limitations, a shift is proposed that reframes the structural analysis as a code generation task. Accordingly, an LLM-empowered agent is developed that (a) integrates chain-of-thought and few-shot prompting to generate accurate OpeeSeesPy code, and (b) automatically executes the code to produce structural analysis results. Experimental results demonstrate that the agent achieves accuracy exceeding 99.0\% on the benchmark dataset, exhibiting reliable and robust performance across diverse conditions. Ablation studies highlight the complete example and function usage examples as the primary contributors to the agent’s enhanced performance.

\end{abstract}

% keywords can be removed
\begin{quote}
\end{quote}

\section{Introduction}
Large language models (LLMs) have recently emerged as transformative tools in artificial intelligence, revolutionizing how machines comprehend, process and generate human language. State-of-the-art models, including closed-source model GPT-4o \citep{openai2024gpt4o} and their open-source counterparts like LLaMA 3.3 \citep{grattafiori2024llama} and DeepSeek-R1 \citep{guo2025deepseek}, have demonstrate exceptional capabilities across a wide range of open-domain tasks, including language translation \citep{zhu2023multilingual, lu2024llamax}, text summarization  \citep{yang2023exploring, zhang2024benchmarking}, and mathematical reasoning \citep{imani2023mathprompter, ahn2024large}. These advancements have spurred growing interest in specialized, domain-specific applications of LLMs, yielding early promising use cases across fields such as medicine \citep{thirunavukarasu2023large, arora2023promise}, law \citep{kim2024legal, graham2023natural}, and finance \citep{lopez2023can, zhang2023enhancing}. These advances have been achieved not by training models from scratch, but by developing effective adaptation techniques that repurpose LLMs for domain-specific tasks \citep{shen2024tag, wei2024drugrealign}. These techniques can be categorized into parameter-free methods and parameter-update methods. Parameter-free methods utilize prompt engineering strategies, such as chain-of-thought prompting \citep{wei2022chain}, few-shot prompting \citep{song2023llm, ji2024zero}, and prompt chaining \citep{wu2022ai}. These strategies allow LLMs to learn task formats from expert workflows or limited examples without retraining. Retrieval-augmented generation (RAG) further enhances these methods by incorporating external knowledge bases during inference \citep{gao2023retrieval}. In contrast, parameter-update methods involve modifying partial or full model parameters, including techniques such as full fine-tuning \citep{lv2023full} and parameter-efficient approaches like low-rank adaptation (LoRA) \citep{hu2022lora}. Collectively, these approaches offer great promise in adapting LLMs to domain-specific challenges.

In civil engineering, researchers have also begun to explore the potential of LLMs for a range of specialized applications. One prominent direction involves natural language processing for engineering documentation. For instance, \cite{zhong2024domain} established a construction management corpus and fine-tuned LLMs to support automated compliance checking and asset condition prediction. Similarly, \cite{joffe2025framework} developed a retrieval-augmented LLM-based chatbot to answer user queries regarding building codes and standards. Another line of work leverages LLMs to automate code generation for scientific computing applications. For example, \cite{kim2025can} assessed the capability of ChatGPT to generate finite element code for simulating hydro-mechanically coupled processes. \cite{pandey2025openfoamgpt} developed an LLM agent capable of automatically generating and executing code for computational fluid dynamics (CFD) simulations. Beyond textual and coding tasks, LLMs have also been applied to visual data such as synthesizing geological cross-sections from sparse site investigation data \citep{li2025few, qian2025large} and monitoring construction equipment from video frames \citep{jeoung2025zero}. Despite these advancements, hallucination remains a critical challenge in LLMs \citep{huang2025survey, tonmoy2024comprehensive}. This issue is particularly important in structural engineering, where the consequences for errors can be substantially larger than in other fields. Consequently, a fundamental yet unanswered question arises: \textit{how reliable and robust do LLMs perform in structural analysis?}

This study aims to take the first step to answer this question by evaluating the performance of a state-of-the-art LLM on simple structural analysis problems that require rigorous understanding of structural mechanics and precise numerical computation. To ensure open-source availability, computational feasibility, and strong instruction-following capabilities, a lightweight LLM, i.e., Llama-3.3 70B Instruct model is selected. A benchmark dataset comprising eight representative beam analysis problems is constructed to assess the model’s performance. Results indicate that while LLaMA-3.3 exhibits a qualitative understanding of structural mechanics, it lacks the reliability (accuracy of correct outputs under repetitive runs of the same analysis problems) and robustness (adaptability to diverse load and boundary conditions) required for engineering applications. To address these limitations, a change of approach is proposed to reframe structural analysis from open-ended text generation tasks to structured code generation tasks. Accordingly, an LLM-empowered agent is developed to integrate chain-of-thought and few-shot prompting techniques to guide LLMs in generating accurate OpenSeesPy \citep{zhu2018openseespy} code. The generated code is then automatically executed by the agent to provide structural analysis results. Experimental results show that the proposed agent significantly improves the reliability and robustness of LLMs in structural analysis, offering a promising pathway for integrating LLMs into structural engineering practice.

\section{Benchmark}
\label{sec:headings}

\subsection{Dataset overview}
\begin{figure*}[htbp]
\centering
% \captionsetup{justification=centering}
\includegraphics[width=0.8\textwidth]{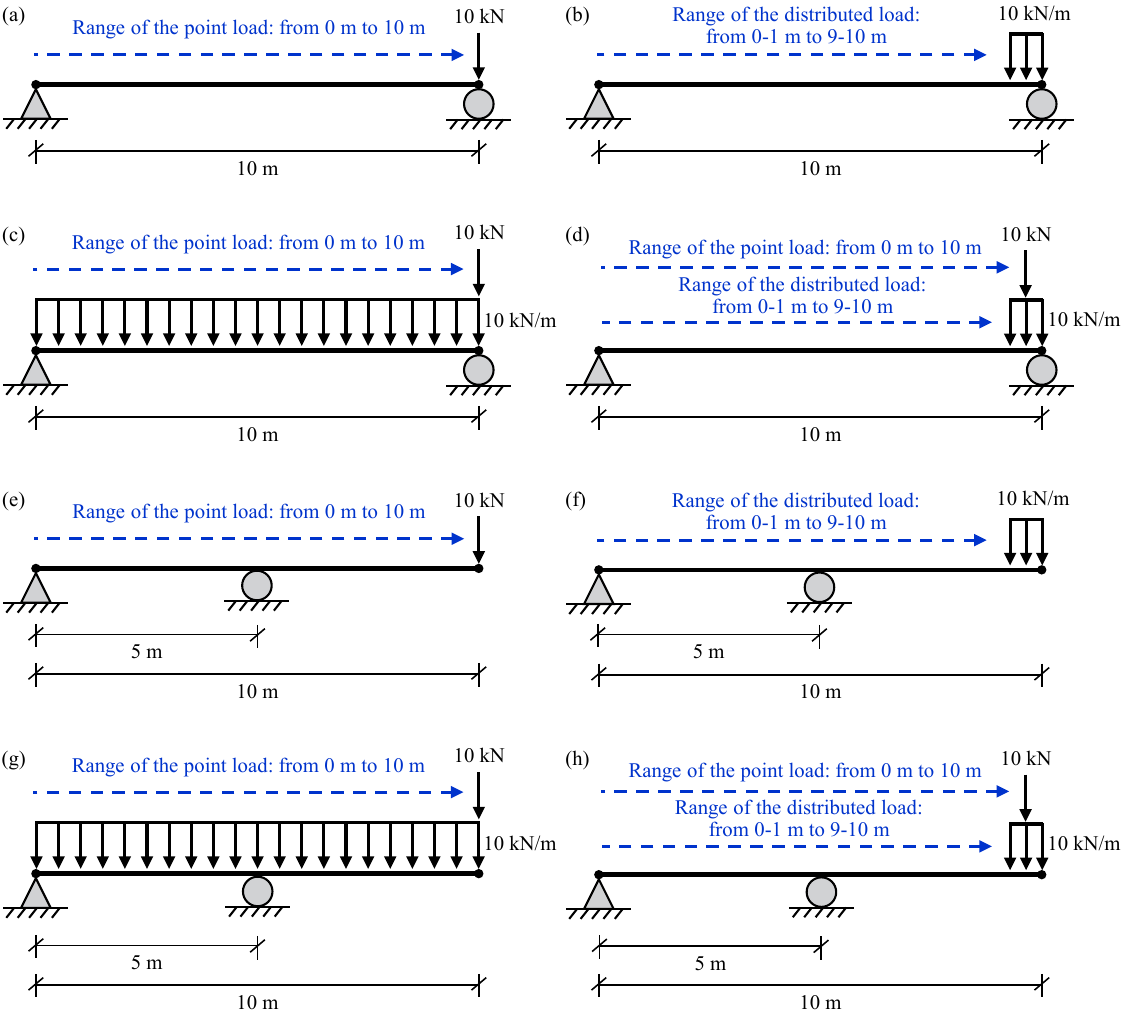}
\caption{A benchmark dataset comprising eight representative beam structural analysis problems.}
\label{Figure1}
\end{figure*}

A benchmark dataset is constructed to evaluate the performance of LLMs on beam structural analysis problems, as illustrated in \cref{Figure1}. Beam structural analysis constitutes a fundamental component of structural engineering as it involves foundational concepts such as support conditions, load distribution, and static equilibrium. These concepts form the basis for understanding how external loads are transferred through structural members and how internal forces, including axial forces, shear forces, and bending moments, evolve along the beam. Proficiency in these analyses is essential for civil engineers as it underpins the design and evaluation of more complex structural systems. Herein, as an first step in evaluating the performance of LLMs in structural engineering, the proposed benchmark focuses specifically on the static equilibrium analysis of beams. The primary objective of the benchmark is to solve the reaction forces at the supports of statically determinate beams, including their magnitudes and directions.   

To comprehensively assess the performance of LLMs in structural analysis, the benchmark comprises eight representative beam problems incorporating diverse load and boundary conditions. Specifically, two canonical beam types are included: simply supported beams and overhang beams, each with a span of 10 meters. In the case of overhang beams, the roller support is positioned at the mid-span. For each beam type, four distinct load conditions are evaluated: (\RN{1}) a 10 kN point load, (\RN{2}) a 10 kN/m uniformly distributed load over a 1-meter segment, (\RN{3}) a combination of a 10 kN point load and a full-span 10 kN/m uniformly distributed load, and (\RN{4}) a combination of a 10 kN point load and a 1-meter 10 kN/m uniformly distributed load. These loads are applied at varying locations along the beam span to reflect a range of realistic design scenarios.

\subsection{Workflows to assess reliabity and robustness}
\begin{figure*}[htbp]
\centering
% \captionsetup{justification=centering}
\includegraphics[width=0.8\textwidth]{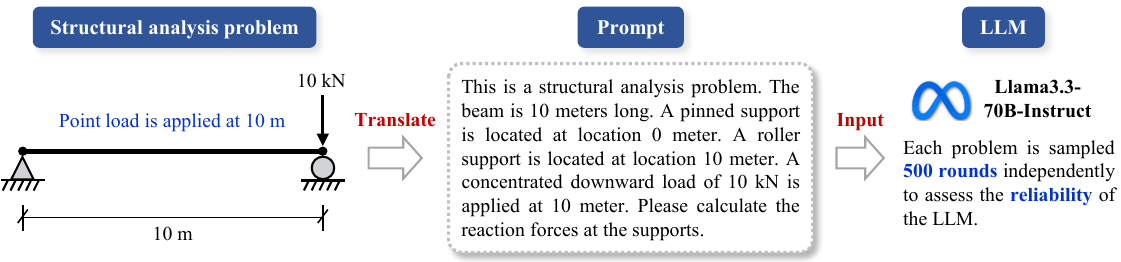}
\caption{Workflow to assess the reliability of LLMs for structural analysis problems.}
\label{Figure2}
\end{figure*}

The assessment of LLM performance in structural analysis focuses on two key dimensions: reliability and robustness, both of which are essential to ensuring that LLMs can serve as trustworthy and confidently deployable tools in engineering practice. Specifically, reliability refers to the accuracy of correct outputs across multiple independent runs of the same analysis problem. This property is particularly important in structural engineering, where even minor inconsistencies can undermine safety and lead to severe consequences. To assess reliability, a standardized workflow is developed, as shown in \cref{Figure2}. For each beam structural analysis problem, a textual prompt is provided to specify the beam geometry, support conditions, load conditions, and target outputs. Recognizing the sensitivity of LLM performance to prompt formulation, a series of prompt variants are designed and tested in a preliminary step. These variants explore different levels of descriptive detail and structural formatting. The prompt presented here represents the most effective formulation identified through iterative experiments, consistently yielding the highest reliability across various problems. Then, this human-labeled prompt is input into the LLM for 500 independent runs. The reliability of the model is quantified through the proportion of correct responses across these repeated runs, as defined in Eq.~(\ref{eq:reliability}):
\begin{equation}
\label{eq:reliability}
\textnormal{Reliability} = \frac{N_{\textnormal{correct}}}{N_{\textnormal{total}}}
\end{equation}
where \( N_{\textnormal{correct}} \) represents the number of correct outputs and \( N_{\textnormal{total}} \) denotes the total number of independent runs, i.e., 500 runs.

\begin{figure*}[htbp]
\centering
% \captionsetup{justification=centering}
\includegraphics[width=0.8\textwidth]{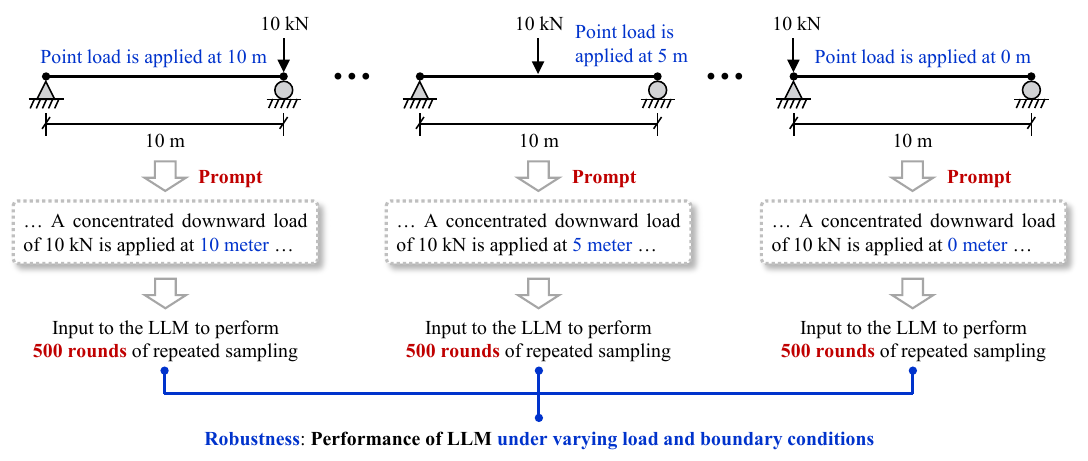}
\caption{Workflow to assess the robustness of LLMs for structural analysis problems.}
\label{Figure3}
\end{figure*}

Robustness is defined as the model’s ability to maintain its performance under variations in input parameters \citep{yu2025benchmarking}. Herein, the robustness is evaluated with respect to reliability and the variations refer to the changes in load and support conditions. This attribute is vital for ensuring the model’s adaptability to a wide range of real-world scenarios. To assess robustness, each structural analysis problem in the benchmark dataset (\cref{Figure1}) is evaluated under a series of varied load conditions, wherein the applied load is incrementally moved from the right end to the left end of the beam in 1-meter intervals, as demonstrated in \cref{Figure3}. Each variation is studied independently, and for each variation, a corresponding textual prompt is provided and submitted to the LLM for 500 independent runs. To ensure a bounded and interpretable robustness metric, a normalized robustness score is defined with reference to the inverse transformation of the coefficient of variation (CV) \citep{hoes2009user}, as defined in Eq.~(\ref{eq:robustness}):
\begin{equation}
\label{eq:robustness}
\mathrm{Robustness} = \left(1 + \mathrm{CV} \right)^{-1} = \left(1 + \frac{\sigma_R}{\bar{R}}\right)^{-1}
\end{equation}
where \( \bar{R} \) and \( \sigma_R \) denote the mean and standard deviation of the reliability under varying load conditions, respectively. This formulation ensures that the robustness value lies within the interval \( (0, 1] \), where higher values indicate stronger robustness.

\subsection{Performance of Llama-3.3 70B Instruct model}

\begin{figure*}[htbp]
\centering
% \captionsetup{justification=centering}
\includegraphics[width=0.8\textwidth]{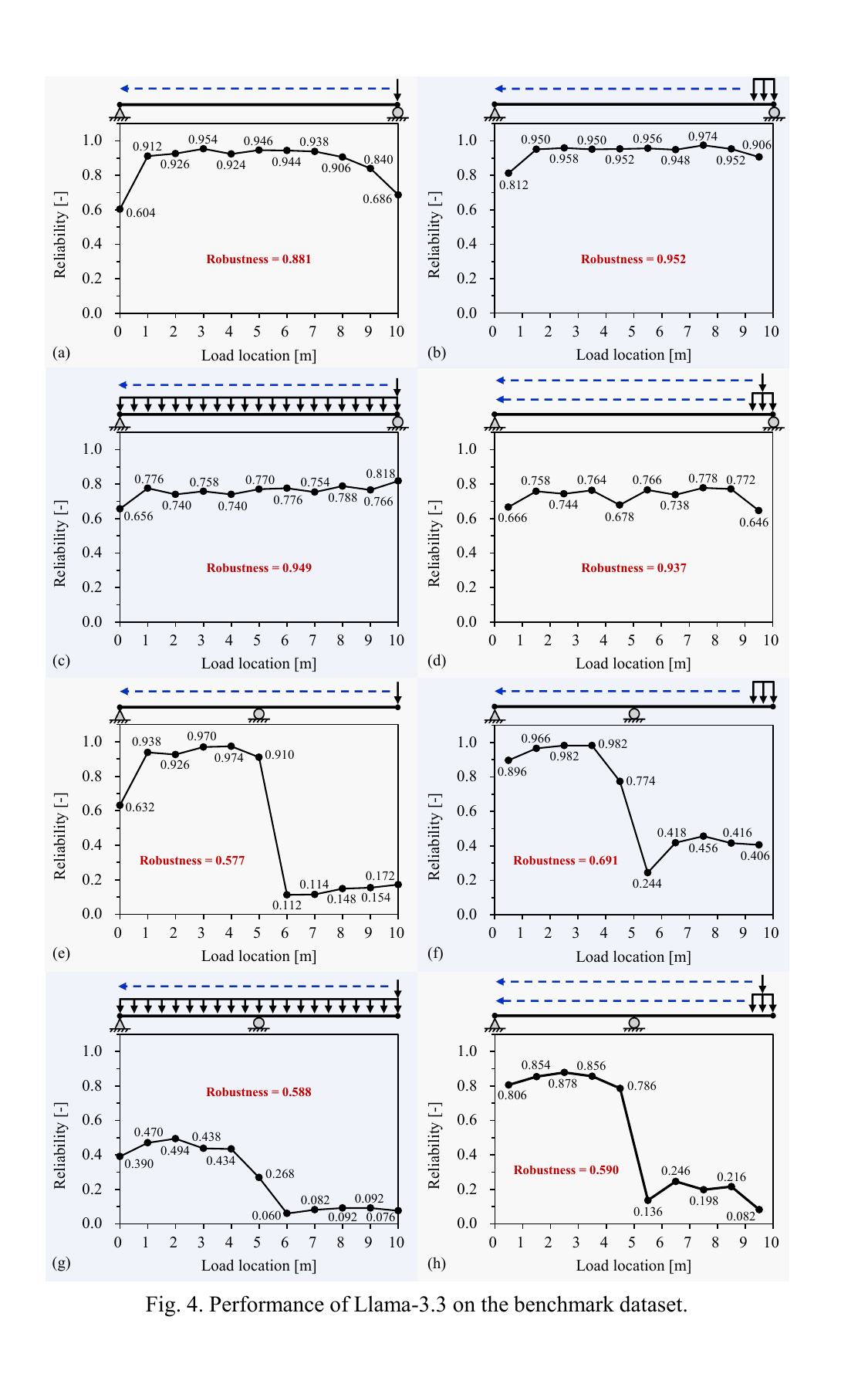}
\caption{Reliability and robustness of Llama-3.3 70B Instruct model on the benchmark dataset.}
\label{Figure4}
\end{figure*}

Following the evaluation workflows described above, the reliability and robustness of the Llama-3.3 70B Instruct model are assessed using the benchmark dataset. The results are presented in \cref{Figure4}, where each subplot corresponds to a specific combination of load and boundary conditions. In each case, the load is applied at different locations along the beam span, producing a reliability curve. Each point on the curve represents the reliability of the model at a given load location, computed from 500 independent runs. Specifically, several key patterns are observed in the results of simply supported beams: (\RN{1}) The LLM demonstrates strong performance under single-load scenarios, achieving average reliability values of 87.1\% for point loads and 93.6\% for localized distributed loads, as shown in \cref{Figure4} (a) and (b). (\RN{2}) The performance of LLM degrades notably as the load conditions become increasingly complex. When a point load is combined with a full-length distributed load or a localized distributed load, the average reliability drops to 75.8\% and 73.1\%, respectively, as depicted in \cref{Figure4} (c) and (d). (\RN{3}) The LLM exhibits high reliability when loads are applied within the span of the beam. However, its robustness deteriorates significantly when the loads are positioned at the supports, as evidenced by the sharp reliability dips at the beam ends across \cref{Figure4} (a) to (d).

In contrast to the simply supported beams, the overhang beam exhibits a distinctly different reliability pattern. As illustrated in \cref{Figure4} (e) to (h), the reliability of the LLM declines dramatically when the load is applied outside the region bounded by the supports, i.e., in the cantilevered region, compared to when the load is applied within the supported span. Particularly, for the case involving a combination of point loads and full-length distributed loads, the value of reliability even falls below 10.0\%, as shown in \cref{Figure4} (g). Additionally, two patterns observed in simply supported beams are also evident in the overhang beams. (\RN{1}) The average reliability of the LLM decreases with increasing load complexity. This can be demonstrated by comparing the performance between the single-load scenarios in \cref{Figure4} (e) and (f) and the combined load cases in \cref{Figure4} (g) and (h). (\RN{2}) A reduction in reliability is also observed when the load is applied at support locations, as shown by the lower reliability values at the leftmost and mid-span positions across \cref{Figure4} (e) to (h).

\begin{figure*}[htbp]
\centering
% \captionsetup{justification=centering}
\includegraphics[width=0.8\textwidth]{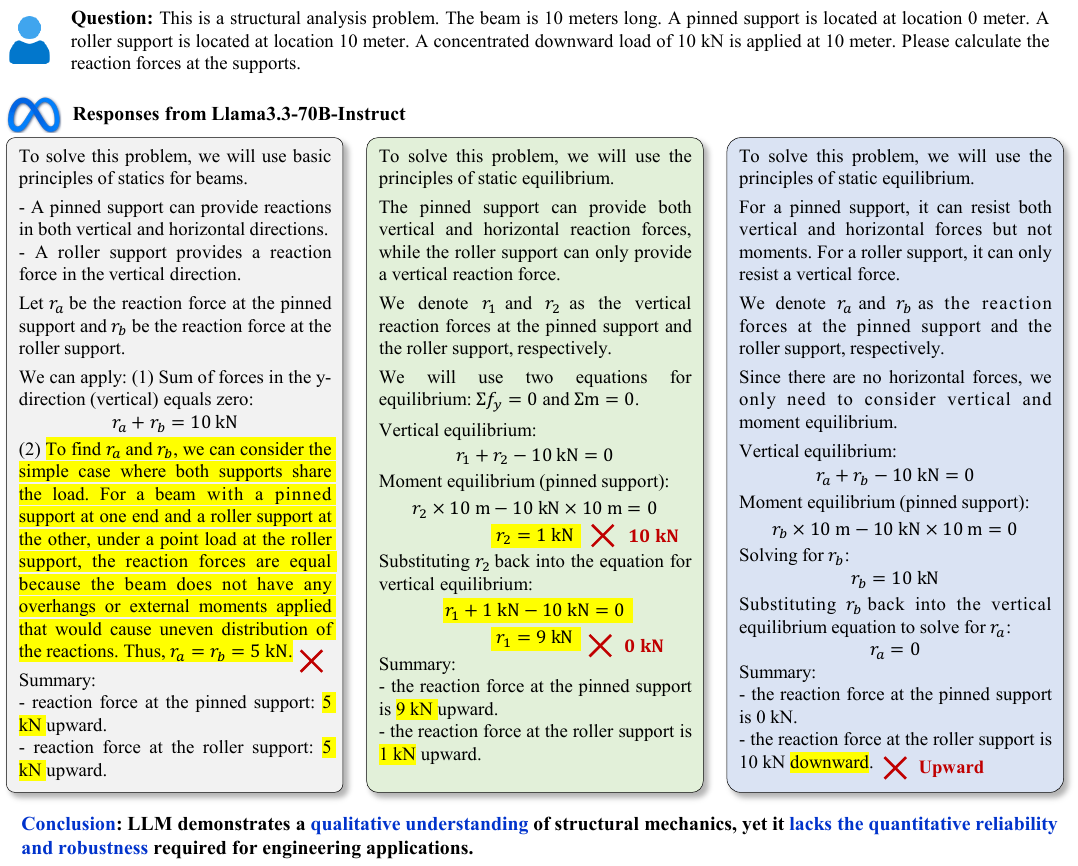}
\caption{Illustrative examples of incorrect solutions produced by the Llama-3.3 70B model in structural analysis tasks.}
\label{Figure5}
\end{figure*}

To qualitatively assess the LLM’s understanding of the fundamental principles underlying structural analysis tasks, three representative incorrect responses generated by the Llama-3.3 70B Instruct model are presented in \cref{Figure5}. In example 1, the LLM correctly identified static equilibrium as the basic principle and formulated the vertical force equilibrium equation. However, it failed to incorporate the moment equilibrium equation and instead relied on an improper assumption that the load was equally shared by the pinned and roller supports. This assumption neglects the fact that the load was applied directly at the roller support, leading to a fundamental error in reasoning and a wrong numerical solution. In example 2, the LLM correctly applied both the vertical force equilibrium and the moment equilibrium equations. Nevertheless, a computational mistake occurred during the algebraic manipulation of equations, resulting in an incorrect numerical answer despite an otherwise sound approach. In example 3, the LLM successfully formulated and solved both the force and moment equilibrium equations, obtaining the correct magnitudes of the support reactions. However, it incorrectly reported the direction of the reaction force as downward instead of upward, indicating a conceptual misunderstanding in the interpretation of mechanical equilibrium. Collectively, these examples indicate that while the LLM demonstrates a qualitative understanding of structural mechanics, it lacks the quantitative reliability and robustness required for practical engineering applications.

\section{A large language model-empowered agent for structural analysis}
\label{sec:others}

\subsection{Reframe the task: From text generation to code generation}

\begin{figure*}[htbp]
\centering
% \captionsetup{justification=centering}
\includegraphics[width=0.8\textwidth]{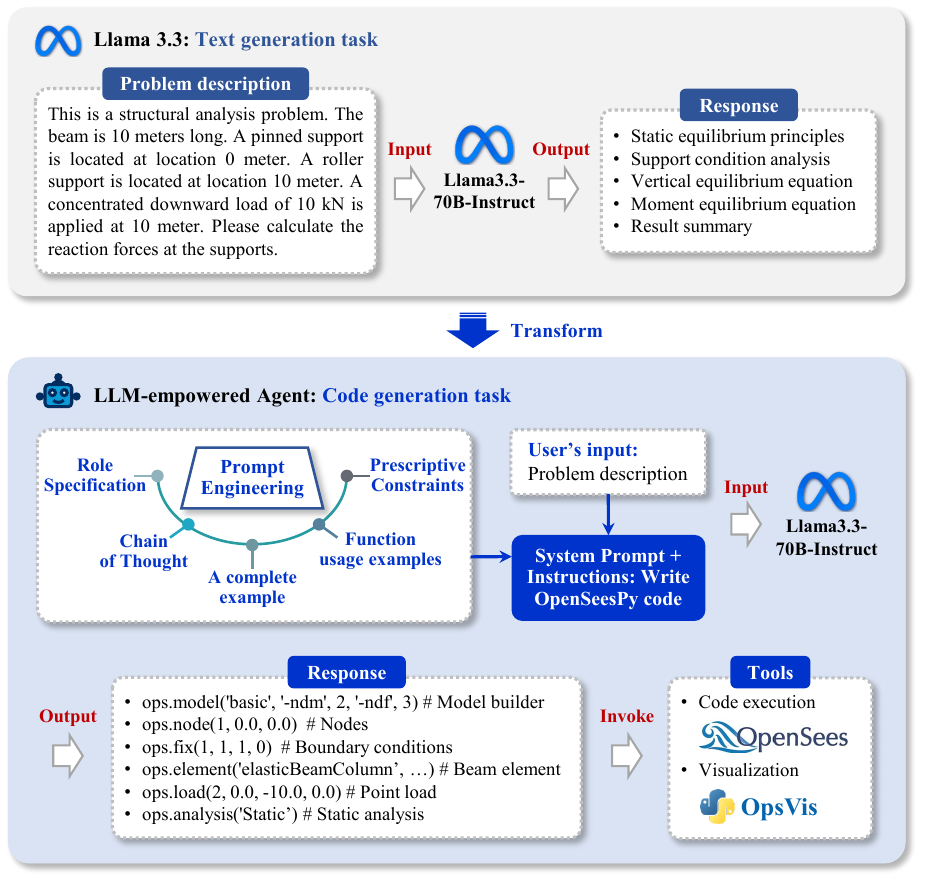}
\caption{A paradigm shift from text generation to code generation for structural analysis tasks using a LLM-empowered agent.}
\label{Figure6}
\end{figure*}

To improve the reliability and robustness of LLM in structural analysis problems, a change of approach is proposed that reframes the task from text generation to code generation, as shown in \cref{Figure6}. This shift is driven by two primary motivations. First, LLM predictions inherently rely on next-token prediction. For text generation tasks, it typically involves high degrees of freedom and uncertainty due to the open-ended nature of natural language. Specifically, natural language is context-dependent and often lacks strict rules, resulting in a vast latent space where a wide range of outputs can be syntactically valid but semantically vague or incorrect. In contrast, code generation operates within a framework defined by rigorous syntactic and semantic rules, which significantly constrains the output space and reduces predictive uncertainty. This structured nature enables LLMs to generate output that conforms to the conventions of programming languages, thereby enhancing consistency and reliability. Existing studies have demonstrated that LLMs tend to exhibit higher accuracy in code generation tasks compared to text generation tasks due to the reduced ambiguity \citep{li2023starcoder, roziere2023code}. Therefore, reframing structural analysis as a code generation task allows LLMs to capitalize on their strengths in modeling structured sequences, thereby producing more reliable and robust outputs for engineering problem-solving.

Second, code generation offers a more scalable approach for adapting LLMs to domain-specific tasks \citep{zhang2024renaissance}. Rather than requiring LLMs to internalize specialized domain knowledge and perform complex symbolic computations, which may lead to hallucinated or incorrect output due to their probabilistic nature, it is more effective to integrate LLMs with established computational tools. This strategy aligns with current practices in civil engineering, where professionals routinely rely on dedicated software platforms to address real-world problems. By positioning the LLM as a language-to-action bridge, the proposed paradigm synergistically leverages the superior capability of LLMs in code generation and the reliability and robustness of domain-specific computational software. Such integration is expected to reduce the cognitive burden on users and promote the automation of task analysis and execution. Additionally, the modularity and reusability of codes facilitate a scalable workflow, enabling LLMs to break down complex, high-level problems into smaller, programmable subtasks. This capacity can enhance the adaptability and practical utility of LLMs in practical engineering applications.

\subsection{Agent architecture}

To enable the automated generation and execution of structural analysis codes, an LLM-empowered agent is developed, specifically designed to integrate with the OpenSeesPy platform. OpenseesPy is an open-source Python library that supports finite element modeling for a variety of structural types, including beams, trusses, and frames \citep{zhu2018openseespy}. For visualization, OpenSeesPy is integrated with the OpsVis package \citep{kokot_opsvis_2024}, which provides graphical representations of structural behavior, thereby enhancing result interpretability and user interaction through intuitive visual feedback.

The architecture of the agent incorporates a dedicated prompt engineering layer designed to enhance the capabilities of LLMs in generating accurate and domain-specific OpenSeesPy code, as illustrated in \cref{Figure6}. This layer comprises five components. The predefined prompts are integrated with the user-provided input, typically a natural language problem description, to construct the system prompt. The system prompt is then input into the LLM, along with explicit instructions to generate OpenSeesPy code corresponding to the described structural analysis task. Upon generating the code, the agent proceeds to invoke OpenSeesPy for numerical simulation and OpsVis for result visualization. This modular workflow allows each component to operate in a well-defined and interpretable pipeline. By leveraging the strengths of both LLMs in code generation and domain-specific computational solvers in structural analysis, the agent provides a reliable and interpretable framework for automating structural analysis.

\subsection{Prompt design}

\begin{figure*}[htbp]
\centering
% \captionsetup{justification=centering}
\includegraphics[width=0.8\textwidth]{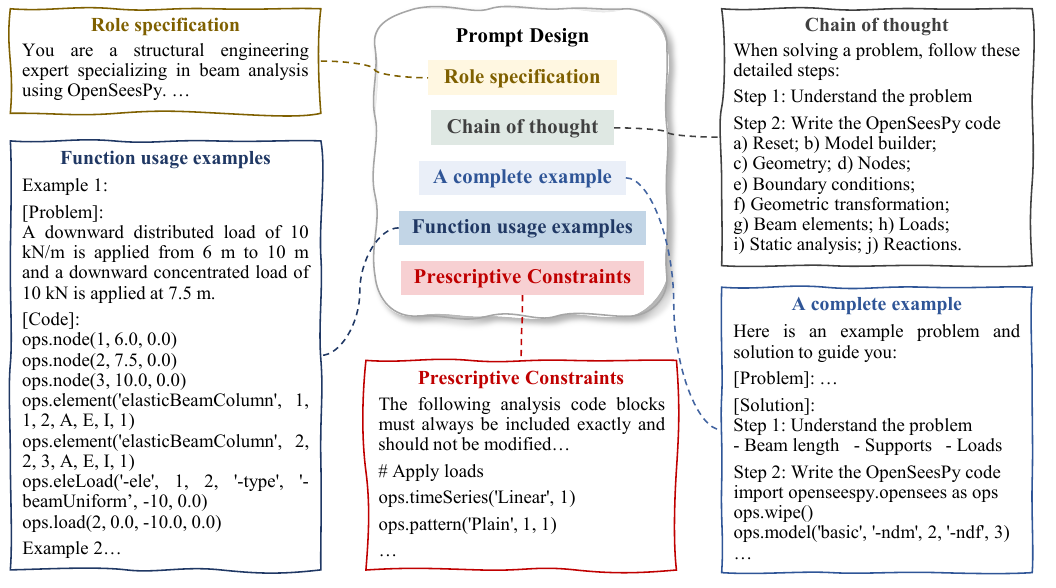}
\caption{A prompt template for reliable and robust structural analysis.}
\label{Figure7}
\end{figure*}

To improve the accuracy, reliability, and domain alignment of LLM outputs in beam analysis problems, a structured prompt template is proposed, as shown in \cref{Figure7}. This template consists of five key components: role specification, chain of thought, a complete example, function usage examples, and prescriptive constraints. The overarching design philosophy leverages two complementary prompting strategies: chain of thought reasoning and few-shot learning. The former encourages the LLM to reason through intermediate steps, decomposing complex structural problems into manageable sub-tasks. The latter provides concrete examples, ranging from complete code generation sequences to modular snippets for key functions, to ground the LLM’s responses in domain-relevant logic and syntax. This dual mechanism mitigates semantic drift, enhances output consistency, and reduces the risk of hallucination, thereby facilitating the transformation of the LLM from a general-purpose language model into a task-specialized assistant for structural analysis.

Each component in the prompt template serves as a distinct role in guiding the generative behavior of the LLM. Specifically, the role specification component assigns a clear identity to the LLM, instructing it to act as a structural engineering expert with specialized knowledge in beam analysis using OpenSeesPy. This ensures that the model activates relevant domain knowledge and maintains a consistent tone and context throughout the response. The chain of thought is divided into two stages. First, the LLM is prompted to interpret the problem by identifying key aspects including beam geometry, boundary conditions, loading conditions, and analysis objectives. Second, the LLM is prompted to outline the logical sequence for code construction, emulating the reasoning process of a human engineer. Following the reasoning step, an example is introduced to provide a full instance of code generation for a simple beam problem. Note that this example demonstrates the complete workflow, including chain of thought reasoning and executable code writing. It serves as a reference from which the LLM can learn the logic and structure for similar tasks.

To further enhance task-specific performance, a set of function usage examples are incorporated to address common weakness in LLM-generated outputs. These challenges primarily involve the definition of nodes, elements, and load cases, particularly under complex loading conditions such as concentrated and distributed loads applied concurrently. The function usage examples begin with instructional tutorials on critical OpenSeesPy functions, providing detailed explanations of input parameters related to beam elements, boundary conditions, and load applications. This is followed by three applied examples that demonstrate how to construct models in which a concentrated load lies within the span of a distributed load. These examples help the LLM learn to correctly identify nodes, segment beam elements, and apply complex loading configurations in a robust and generalizable manner. Finally, prescriptive constraints are incorporated to regulate code formatting, where mandatory code blocks are specified to prevent unauthorized modifications. These constraints enhance code compatibility with downstream analysis tools and minimize the risk of execution errors.

\section{Results and discussion}

\subsection{Performance of the proposed LLM-empowered agent}

\begin{figure*}[htbp]
\centering
% \captionsetup{justification=centering}
\includegraphics[width=0.8\textwidth]{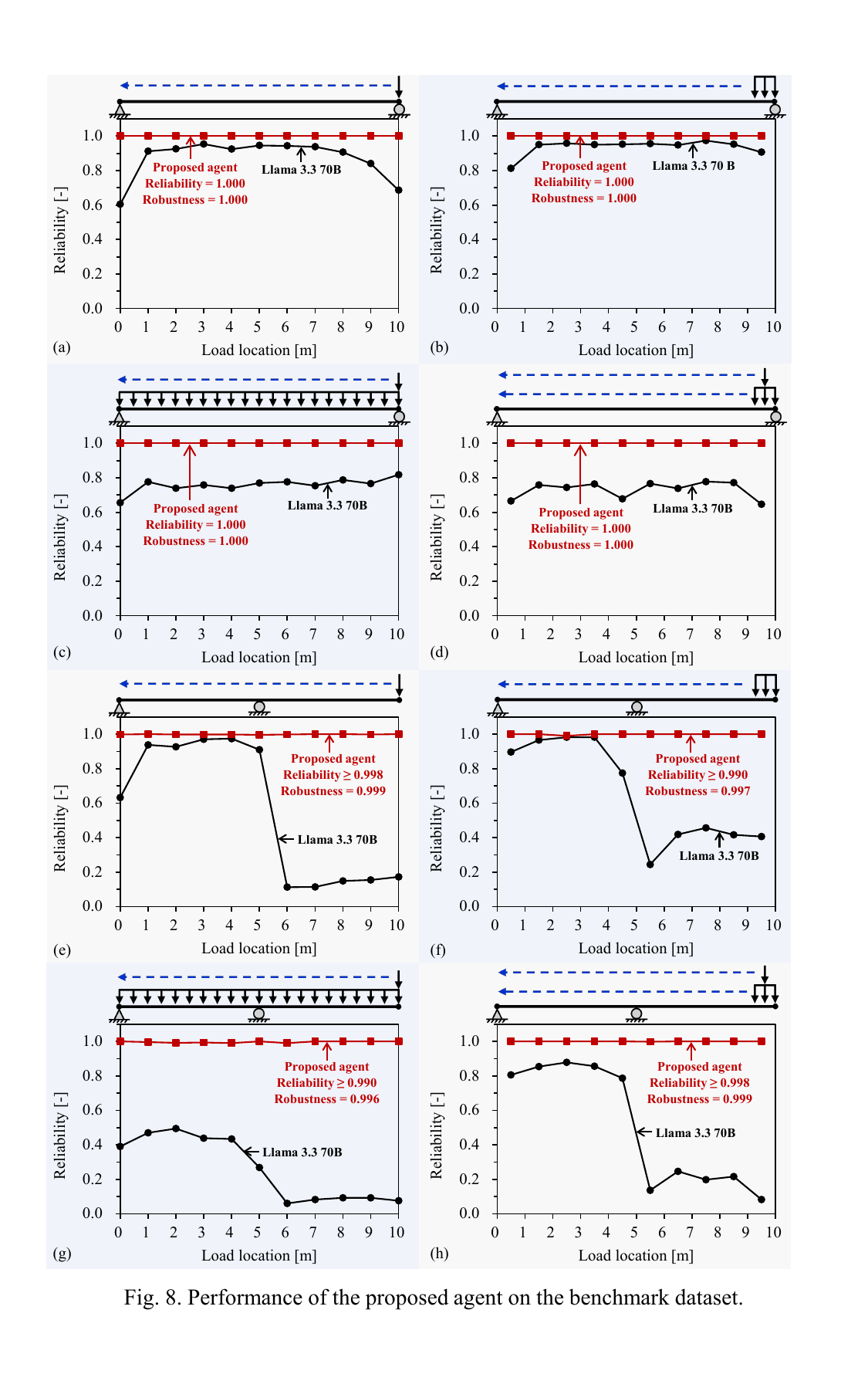}
\caption{Reliability and robustness of the proposed LLM-empowered agent on the benchmark dataset.}
\label{Figure8}
\end{figure*}

The performance of the proposed LLM-empowered agent is assessed using the benchmark dataset in terms of reliability and robustness. As illustrated in \cref{Figure8}, the proposed agent consistently achieves reliability over 0.990 and robustness over 0.996 across a range of load and boundary conditions, significantly outperforming the baseline Llama-3.3 70B Instruct model. Particularly, the agent exhibits substantial improvements in reliability under conditions where the baseline model’s performance degrades, such as under point loads or localized distributed loads applied at support locations or in cantilevered regions. These improvements yield flat reliability curves, demonstrating the strong robustness of the proposed agent in structural analysis tasks. The enhanced performance can be attributed to two key strategies. First, by reframing structural analysis as a code-generation task, the problem is transformed into a structured and standardized workflow. This mitigates ambiguity or inaccuracy in natural language reasoning, thereby improving the robustness of the agent across varied conditions. Second, the incorporation of dedicated prompt templates provides reasoning guidance and concrete examples, which enhances the consistency and reliability of the agent’s outputs over repetitive runs. Together, these strategies enable a reliable and robust LLM-empowered agent for automated structural analysis. Several representative execution examples of the proposed agent are provided in \hyperref[appendix:examples]{Appendix}.

Additionally, a detailed analysis of the agent-generated incorrect cases is conducted, identifying two primary error types. The first type involves the misapplication of boundary conditions in cantilever beam problems. For these cases, the correct configuration involves a pinned support at the left end and a roller support at the mid-span, leaving the right end as a free cantilever. However, the agent occasionally hallucinates an additional roller support at this free end. This is a typical form of LLM hallucination, where the model defaults to more common structural configurations learned during its training, rather than strictly adhering to the prescribed conditions in the problem statement. The second error type involves the inaccurate application of loads, particularly under combined load conditions. For example, when tasked with applying a distributed load from 1 to 2 meters and a point load at 1.5 meters, the agent extends the range of the distributed load to 0 to 2 meters. Although function usage examples have been demonstrated in the prompt template, the agent occasionally struggles to precisely parse the combined load conditions, leading to intermittent inaccuracies in load application.

\subsection{Generalization of the performance analysis on extended beam tasks}

\begin{figure*}[htbp]
\centering
% \captionsetup{justification=centering}
\includegraphics[width=0.4\textwidth]{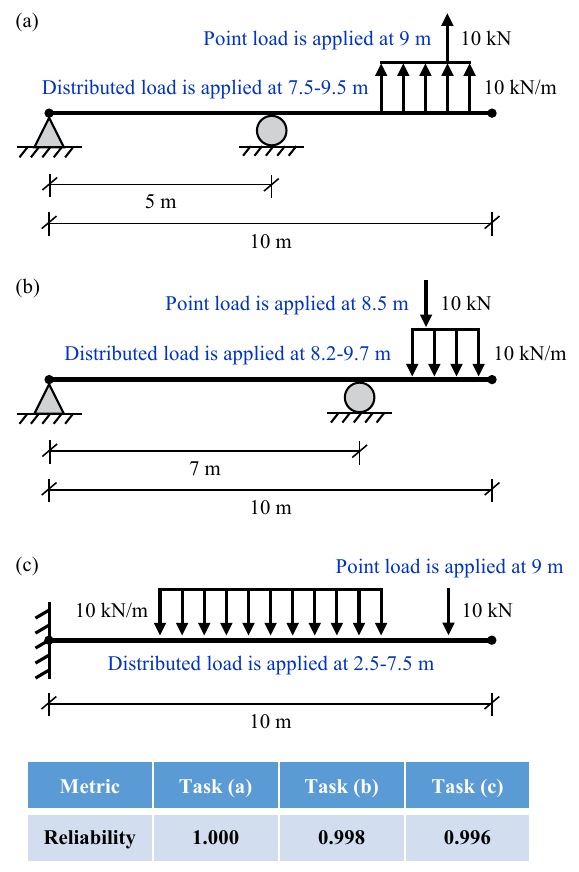}
\caption{Generalization performance of the proposed LLM-empowered agent on extended beam analysis tasks.}
\label{Figure9}
\end{figure*}

Beyond the benchmark dataset, the proposed LLM-empowered agent is further tested on a series of more complicated beam analysis problems to assess its generalization capabilities and robustness under varied structural conditions. As illustrated in \cref{Figure9}, these test cases introduce variations in load locations, load directions, support types, and support locations, all of which differ from the conditions presented in the benchmark dataset. Specifically, three representative problems are designed. (a) An overhang beam with a pinned support at 0 meters and a roller support at 5 meters. The beam is subjected to two upward loads: a point load at 9 meters and a uniformly distributed load spanning from 7.5 to 9.5 meters. (b) An overhang beam with a pinned support at 0 meters and a roller support at 7 meters. The beam is subjected to two downward loads: a point load at 8.5 meters and a uniformly distributed load ranging from 8.2 to 9.7 meters. (c) A cantilever beam with a fixed support at 0 meters. The beam is subjected to two downward loads: a point load at 9 meters and a uniformly distributed load spanning from 2.5 to 7.5 meters.

For each test case, a problem description in natural language is provided to the agent as input. In problems (a) and (b), the agent is tasked with solving the reaction forces at the pinned and roller supports, whereas in problem (c), the agent is instructed to determine both the reaction forces and bending moment at the fixed support. To assess the reliability of the agent under these generalized conditions, each problem is executed independently over 500 runs. The results indicate that the proposed agent maintains strong reliability across three extended scenarios, achieving reliability of 100\%, 99.8\%, and 99.6\% for problems (a), (b), and (c), respectively. These findings highlight the agent’s capability to generalize to novel beam analysis scenarios while maintaining reliable and robust performance. Such generalization ability is essential for real-world applications in structural engineering, where each structure may present distinct loading and boundary constraints.

\begin{figure*}[htbp]
\centering
% \captionsetup{justification=centering}
\includegraphics[width=0.8\textwidth]{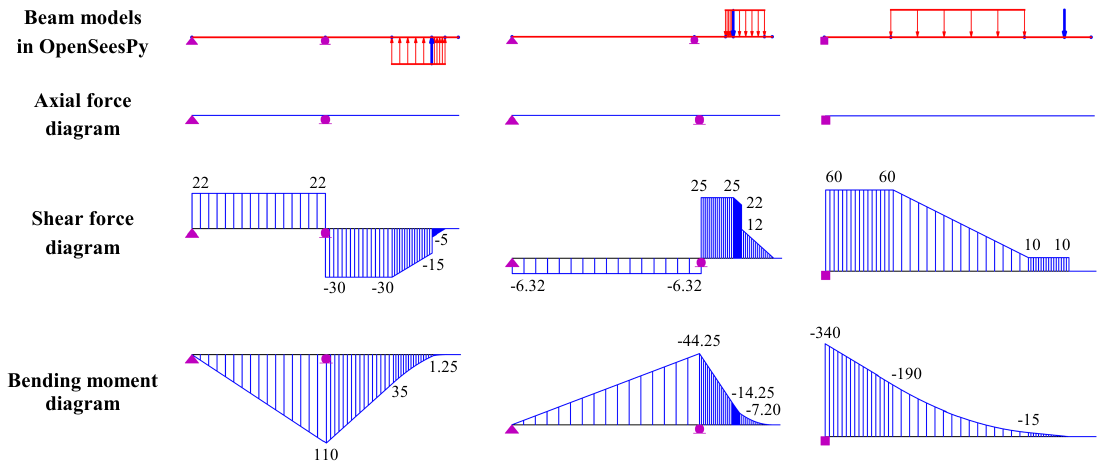}
\caption{Visualization of the proposed LLM-empowered agent for extended beam analysis problems.}
\label{Figure10}
\end{figure*}

OpenSeesPy codes produced by the LLM-empowered agent are automatically executed to compute the reaction forces and bending moments. The resulting structural responses are subsequently visualized using the OpsVis package, as shown in \cref{Figure10}. These visualizations include the finite element models along with the corresponding axial force, shear force, and bending moment diagrams for each case. The constructed finite element models exhibit strong alignment with domain-specific physical expectations, accurately reflecting the intended node configurations, loading conditions, and support placements. This consistency confirms that the proposed agent generates syntactically correct and valid codes. In summary, the visualization module in the agent can significantly enhance the practical utility of the proposed agent by providing engineers with graphical representations of structural behavior, facilitating model validation, error checking, and decision-making in real-world applications. Moreover, the graphical output goes in the direction of providing an explanation for the results, and caters to the desire of explainability in the use of LLMs.

\subsection{Ablation experiments on prompt components}

To systematically assess the contribution of individual components in the structured prompt template, a series of ablation experiments are conducted using the extended beam analysis tasks. In each experiment, one component is removed from the prompt template to isolate its effect, resulting in four prompt configurations: without chain of thought reasoning, without the complete example, without function usage examples, and without prescriptive constraints. Each configuration is applied to all three extended tasks and the reliability of the agent is assessed over 500 independent runs.

The results of these ablation experiments are presented in Table~\ref{tab:ablation}. It is shown that the complete example emerges as the most influential component. Its removal causes the reliability of the agent to dramatically drop to 0\% across all three extended tasks. This behavior is attributed to LLM’s inability to infer the full modeling workflow, resulting in missing or disordered code segments and subsequent execution errors. Function usage examples also play a vital role in maintaining the agent’s performance. When these examples are excluded, the reliability of the agent declines significantly to 2.6\%, 29.6\%, and 62.3\% for the three tasks, respectively. A detailed inspection of the outputs reveals that most errors stem from incorrect node placement and improper application of loads. These findings highlight that the function usage examples improve the LLM’s ability to learn domain-specific modeling strategies under complex load conditions, thereby enhancing the agent’s generalization capability.

\begin{table}[htbp]
\centering
\caption{Ablation experiments on prompt components.}
\label{tab:ablation}
\begin{tabular}{lcccc}
\toprule
\textnormal{Prompt configuration} & \textnormal{Reliability: Task (a)\textsuperscript{*}} & \textnormal{Reliability: Task (b)\textsuperscript{*}} & \textnormal{Reliability: Task (c)\textsuperscript{*}} & \textnormal{Robustness} \\
\midrule
Proposed prompt template         & 1.000          & 0.998          & 0.996          & 0.998\\
w/o chain of thought             & 1.000          & 0.998          & 0.784          & 0.882\\
w/o complete example             & 0.000          & 0.000          & 0.000          & --\\
w/o function usage examples      & 0.026          & 0.296          & 0.623          & 0.513\\
w/o constraints                  & 0.936          & 0.970          & 0.556          & 0.781\\
\midrule
\multicolumn{5}{l}{\textsuperscript{*}Tasks (a), (b), and (c) correspond to extended beam analysis problems shown in \cref{Figure9}.} \\
\end{tabular}
\end{table}

Additionally, the exclusion of the chain of thought or prescriptive constraints leads to more moderate performance degradation, as illustrated in Table~\ref{tab:ablation}. Specifically, without the chain of thought component, the agent still achieves high reliability of 100\% and 99.8\% in the first two overhang beam tasks. However, its performance decreases to 78.4\% in the cantilever beam case. Given that the cantilever case is not included in the prompt template, this observation suggests that chain of thought reasoning helps activate latent domain knowledge within the LLM, thus improving the generalization and robustness of the agent. Similarly, removing prescriptive constraints leads to reliability of 93.6\%, 97.0\%, and 55.6\% across the three tasks. Errors in this condition are primarily due to the omissions of essential code blocks, emphasizing the role of hard-coded constraints in enforcing structural consistency and minimizing execution failure. Collectively, the above results underscore the critical contribution of all prompt components to the reliable and robust performance of the proposed agent in structural analysis applications.

\section{Summary and conclusions}

This paper represents a first step toward assessing and enhancing the reliability and robustness of large language models (LLMs) in structural engineering, with a focus on the static equilibrium analysis of beams. The assessment results reveal that while the Llama-3.3 70B Instruct model demonstrates a qualitative understanding of structural mechanics, it lacks the quantitative reliability and robustness required for practical engineering applications. To address these limitations, the structural analysis task is shifted from text generation to code generation. Accordingly, an LLM-empowered agent is developed to automatically generate and execute OpenSeesPy and OpsVis codes based on natural language input. It shows that the proposed agent achieves strong reliability exceeding 99.0\% and robustness exceeding 99.6\% across all benchmark problems, significantly outperforming the baseline model.

Beyond these findings, several forward-looking insights emerge from the results:

\begin{itemize}
    \item The ablation experiments reveal the central roles of the complete example and function usage examples in enhancing the agent’s performance. These examples enable the LLM to interpret semantic descriptions, learn logical reasoning, and internalize coding syntax from representative cases. This observation underscores the great potential of generalizing the proposed approach to other tasks. By constructing a comprehensive portfolio of task-specific examples, the capabilities of LLMs can be extended to tackle analogous tasks that share similar logical and code patterns. Such a scalable framework provides a promising pathway toward reliable and robust solutions for a wide range of engineering applications.
    
    \item The agent’s ability to produce visual output not only enhances the interpretability of LLMs but also opens new avenues for interactive human–LLM collaboration. Specifically, the visual output can serve as immediate feedback for analysts, allowing engineers to interactively construct complex structural configurations through a series of incremental modifications, starting from simple base cases. This progressive workflow leverages the domain-specific knowledge of human experts to identify and correct potential errors in LLM-generated codes, thereby improving the usability, adaptability, and practical integration of LLM-based tools in engineering applications.
    
    \item While this study focuses on structured natural language input, the agent occasionally hallucinates additional loads or boundary conditions that are not specified in the problem description. This limitation highlights the potential of employing vision-language models (VLMs), which can jointly process textual descriptions and graphical representations of structures. Additionally, as structural configurations become increasingly complex, such as 2D frames or 3D assemblies, textual descriptions may be difficult to construct and prone to ambiguity. Therefore, leveraging VLMs to interpret both visual and textual inputs may improve model understanding of structural configurations, representing a promising direction for future research and practical deployment.
    
\end{itemize}

\bibliographystyle{ascelike}  
\bibliography{references}  %%% Remove comment to use the external .bib file (using bibtex).
%%% and comment out the ``thebibliography'' section.

%%% Comment out this section when you \bibliography{references} is enabled.
% \begin{thebibliography}{1}

% \bibitem{kour2014real}
% George Kour and Raid Saabne.
% \newblock Real-time segmentation of on-line handwritten arabic script.
% \newblock In {\em Frontiers in Handwriting Recognition (ICFHR), 2014 14th
%   International Conference on}, pages 417--422. IEEE, 2014.

% \end{thebibliography}

\clearpage 
\section*{Appendix: Execution examples of the proposed LLM-empowered agent}
\label{appendix:examples}

\begin{figure*}[htbp]
\centering
% \captionsetup{justification=centering}
\includegraphics[width=0.72\textwidth]{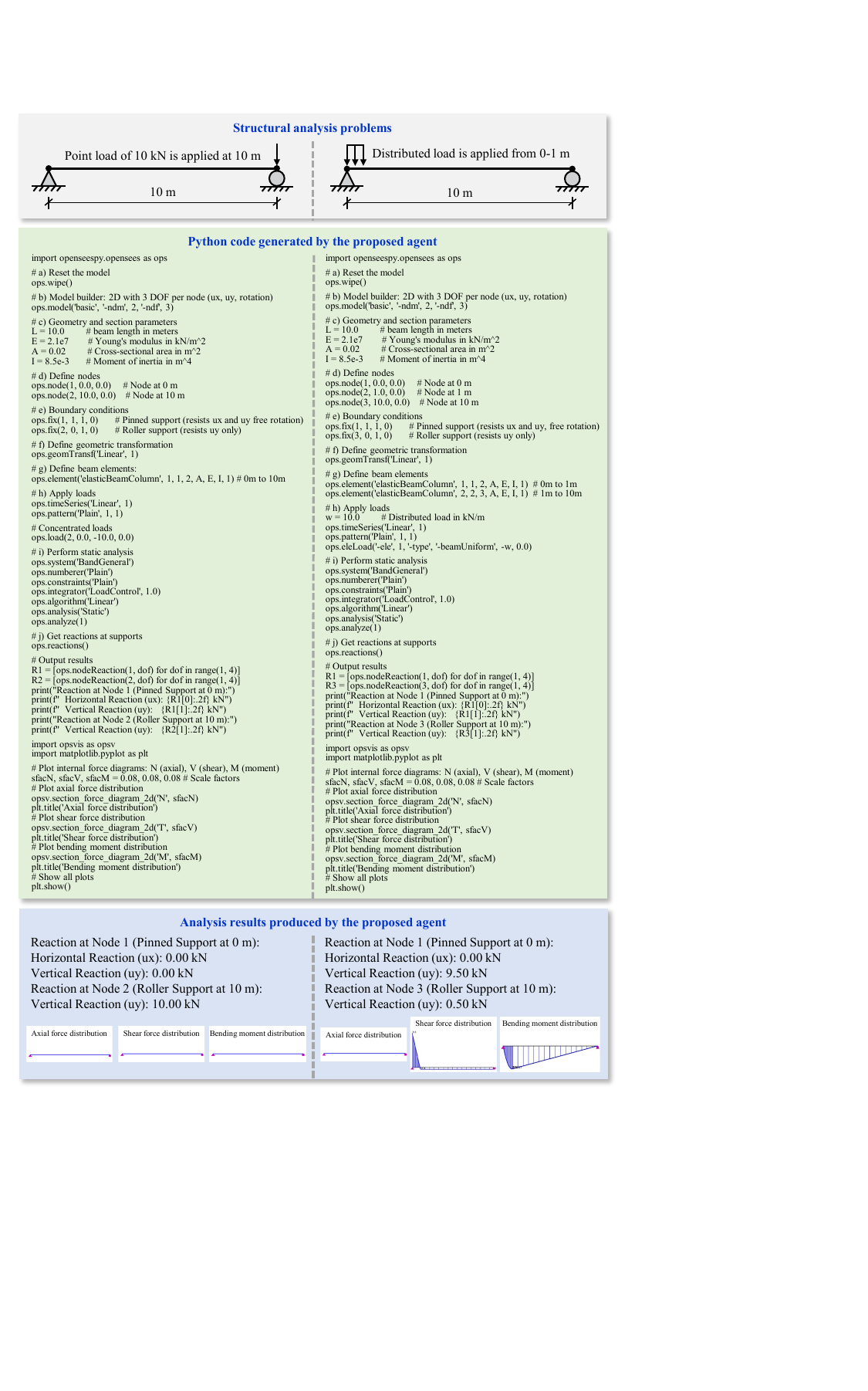}
\caption{Examples of simply supported beam under various load conditions.}
\label{Figure16}
\end{figure*}

\begin{figure*}[htbp]
\centering
% \captionsetup{justification=centering}
\includegraphics[width=0.75\textwidth]{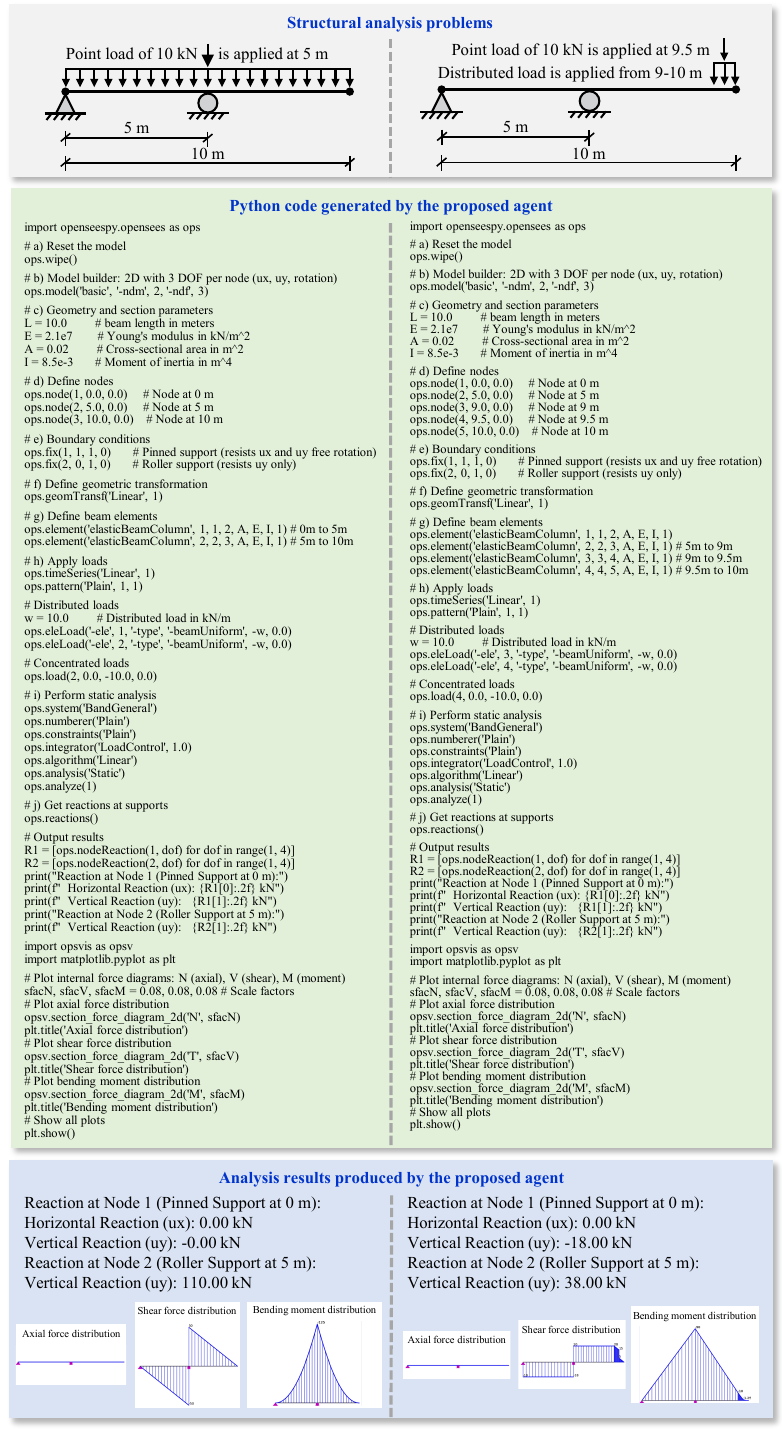}
\caption{Examples of overhang beam under various load conditions.}
\label{Figure17}
\end{figure*}

\end{document}